\documentclass[12pt]{article}

\usepackage{amsthm,amsmath,amssymb}         
\usepackage[round]{natbib}                  
\usepackage{mathabx}                        
\usepackage{tikz, subfigure}                
\usetikzlibrary{arrows.meta}
\usepackage{caption}                        
\usepackage{multirow}                       
\usepackage{booktabs}                       
\usepackage[section]{placeins}              

\RequirePackage[colorlinks,citecolor=blue,urlcolor=blue]{hyperref}

\makeatletter
\def\Ddots{\mathinner{\mkern1mu\raise\p@
\vbox{\kern7\p@\hbox{.}}\mkern2mu
\raise4\p@\hbox{.}\mkern2mu\raise7\p@\hbox{.}\mkern1mu}}
\makeatother

\makeatletter
\def\namedlabel#1#2{\begingroup
	#2%
	\def\@currentlabel{#2}%
	\phantomsection\label{#1}\endgroup
}
\makeatother

\newcommand\independent{\protect\mathpalette{\protect\independenT}{\perp}}
\def\independenT#1#2{\mathrel{\rlap{$#1#2$}\mkern2mu{#1#2}}}

\numberwithin{equation}{section}

\begin{document}

\vskip 3mm

\noindent \textbf{Learning pairwise Markov network structures using correlation neighborhoods}
\vskip 3mm

\vskip 5mm
\noindent Juri Kuronen\textsuperscript{a,b,*}, Jukka Corander\textsuperscript{a,b} and Johan Pensar\textsuperscript{b}

\noindent \textsuperscript{a}Department of Biostatistics, University of Oslo, Norway; \textsuperscript{b}Department of Mathematics and Statistics, University of Helsinki, Finland; \textsuperscript{*}Corresponding author. E-mail address: juri.kuronen@medisin.uio.no

\noindent 

\vskip 3mm

\vskip 5mm

\noindent ABSTRACT

Markov networks are widely studied and used throughout multivariate statistics and computer science. In particular, the problem of learning the structure of Markov networks from data without invoking chordality assumptions in order to retain expressiveness of the model class has been given a considerable attention in the recent literature, where numerous constraint-based or score-based methods have been introduced. Here we develop a new search algorithm for the network score-optimization that has several computational advantages and scales well to high-dimensional data sets. The key observation behind the algorithm is that the neighborhood of a variable can be efficiently captured using local penalized likelihood ratio (PLR) tests by exploiting an exponential decay of correlations across the neighborhood with an increasing graph-theoretic distance from the focus node. The candidate neighborhoods are then processed by a two-stage hill-climbing (HC) algorithm. Our approach, termed fully as PLRHC-BIC$_{0.5}$, compares favorably against the state-of-the-art methods in all our experiments spanning both low- and high-dimensional networks and a wide range of sample sizes. An efficient implementation of PLRHC-BIC$_{0.5}$ is freely available from the URL: \href{https://github.com/jurikuronen/plrhc}{https://github.com/jurikuronen/plrhc}.
\vskip 4mm

\setcounter{page}{1}

\section{Introduction} \label{chapter:introduction}
Markov networks, also known as undirected graphical models, are a popular tool to model high-dimensional probability distributions. The structure of a Markov network, represented by an undirected graph, compactly encodes the conditional independence structure between the variables of the distribution. A set of numerical parameters over this structure then specifies the joint distribution of the model. In this paper, we consider the problem of learning the structure of pairwise Markov networks over binary variables.

Structure learning algorithms can be generally classifed into two broad categories, they either use a constraint-based or a score-based approach to optimize the network topology. In this paper, we focus on the latter approach. Constraint-based algorithms infer the structure through a series of statistical independence tests and the local nature of the tests makes this an attractive approach from the computational scalability perspective. However, a particular drawback is that the individual tests are sensitive to noise which can result in incorrect independence assumptions. In contrast, score-based algorithms operate globally by formulating the structure learning problem as an optimization problem where the aim is to balance between a good fit to the data and avoiding asserting spurious dependencies. This requires a scoring function which measures the level of a penalized fit to data and a strategy to search for a high-scoring structure. There are two main challenges with this approach. Firstly, popular scoring functions based on the likelihood are intractable for larger Markov networks due to a normalizing constant. For this reason, the earliest score-based algorithms were limited to models which constrained the underlying graph to be chordal, since such models allow for a complete factorization of the likelihood (Koller and Friedman \citeyear{koller2009}). The second challenge comes from the exponential growth of the number of possible graph structures in the number of variables which poses a major difficulty to the design of the search algorithm.

Recently, there has been a surge of pseudo-likelihood-based methods which have enabled score-based learning of general, non-chordal Markov network structures. For example, pseudo-likelihood-based model selection using logistic regression has been used to learn Ising models, which are equivalent to binary pairwise Markov networks (Ravikumar, Wainwright and Lafferty \citeyear{ravikumar2010}; Jalali, Johnson and Ravikumar \citeyear{jalali2011}; Barber and Drton \citeyear{barber2015}). Further, Pensar et al. (\citeyear{pensar2017}) used the pseudo-likelihood framework to apply the classical Bayesian-Dirichlet score (Heckerman, Geiger and Chickering \citeyear{heckerman1995}) for learning general Markov network structures. On account of the local Markov property, the pseudo-likelihood allows for deriving tractable and consistent variable-wise scores, which addresses the first challenge. In terms of the search, the scalability of these methods then depends on the ability to break down the global graph discovery problem into a collection of local Markov blanket discovery problems, which can be solved approximately in a reasonable time. However, pseudo-likelihood-based methods still suffer from a considerable computational burden in the search phase because current algorithms require many iterations over the entire data set for each variable.

In this work, following the theoretical work of Bresler, Mossel and Sly (\citeyear{bresler2008}), we introduce a new approach to discover small candidate neighborhoods using only a single iteration over the data set per variable. The idea is based on the correlation decay property that was originally studied in statistical physics models by Dobrushin (\citeyear{dobrushin1970}). As scoring function, we use the extended Bayesian information criterion (BIC$_\gamma$) (Barber and Drton \citeyear{barber2015}), whose penalty term functions as the correlation threshold. Typically, regression-based methods, such as the one by Barber and Drton (\citeyear{barber2015}), require the user to choose between an OR and an AND criterion according to which the final graph is constructed. Choosing between the criteria may be difficult in practice, yet it can have a significant impact on the false positive and negative rates. To avoid this rather arbitrary choice, we follow a similar approach as in Pensar et al. (\citeyear{pensar2017}), where a second global learning phase is applied on the reduced model space. The full approach introduced in this work shows advantage over the state-of-the-art structure learning methods in both speed and learning accuracy in all our experiments.

The remaining article is structured as follows. We review pairwise Markov networks, pseudo-likelihood and the BIC$_\gamma$ score in Section~\ref{section:background}. In Section~\ref{section:search_algorithm}, we review the previously used optimization strategies and introduce our new approach. Section~\ref{section:experiments} contains numerical experiments demonstrating the performance of the compared methods. Section~\ref{section:discussion} concludes the article with a discussion and remarks about possible future research directions.

\section{Structure learning of pairwise Markov networks} \label{section:background}
\subsection{Pairwise Markov networks}
Let $X_V = \{X_1, \ldots, X_d\}$ be a set of $d$ binary random variables with each variable $X_j$ taking values in $\mathcal{X}_j = \{0, 1\}$. Denote the joint outcome space by $\mathcal{X}_V = \bigtimes_{j = 1}^d \mathcal{X}_j$. The dependence structure of a Markov network over $X_V$ is compactly encoded by an undirected graph $G = (V, E)$ where each node in the vertex set $V = \{1, \ldots, d\}$ corresponds to a random variable in the set $X_V$ and the edges $E \subset V \times V$ represent direct dependencies between the variables. In the context of Markov networks, the set of neighbors of node $j$ in graph $G$ is called the Markov blanket of $j$, denoted by $mb(j)$. 

The Markov blanket $mb(j)$ designates for each variable $X_j$ the smallest subset of other variables that when conditioned on make $X_j$ conditionally independent of all the other variables. This is formally expressed as the Local Markov Property:
\begin{description}
    \item[\namedlabel{itm:L}{\normalfont{(L)}}] \emph{Local Markov Property.} $X_j \independent X_{V \setminus \left(j \, \cup \, mb(j)\right)} \mid X_{mb(j)}$ for all $j \in V$.
\end{description}
For positive distributions, the independence properties of the joint probability distribution $p(X_V)$ encoded in graph $G$ imply that $p(X_V)$ factorizes as a product of potential functions over the structure of $G$. In this work, we focus on a special subclass of Markov networks that are restricted to pairwise interactions corresponding to the edges of the graph. The joint distribution of such a network is given by the log-linear parameterization
\begin{equation} \label{eq:joint_pdf}
    p(x_V) = \frac{1}{Z} \exp\left\{\sum\limits_{j \in V} \theta_j x_j + \sum\limits_{(j, j^\prime) \in E} \theta_{j j^\prime} x_j x_{j^\prime}\right\},
\end{equation}
where $\theta_j$ and $\theta_{j j^\prime}$ take real values and $Z$ is a normalizing constant, called the partition function, defined as
\begin{equation} \label{eq:Z}
    Z = \sum\limits_{x_V \in \mathcal{X}_V} \exp\left\{\sum\limits_{j \in V} \theta_j x_j + \sum\limits_{(j, j^\prime) \in E} \theta_{j j^\prime} x_j x_{j^\prime}\right\}.
\end{equation}
The other direction, that the factorization implies the independence properties, is guaranteed by the Hammersley--Clifford theorem (Hammersley and Clifford \citeyear{hammersley1971}). However, it is possible for other independencies in $p(X_V)$ to exist even if they are not represented in $G$. In this work, we make the generally used assumption that $p(X_V)$ is faithful to $G$, which means that $p(X_V)$ does not contain such additional independencies. For a discussion about its implications, see Koller and Friedman (\citeyear{koller2009}, Section 3.3.2).

\subsection{Structure learning using logistic regression}
The likelihood function provides a natural measure for evaluating the fit of a candidate network structure to data. However, maximizing the likelihood involves computing the partition function $Z$ in \eqref{eq:Z}, which is feasible only for the smallest networks. In this work we consider the pseudo-likelihood, introduced originally by Besag (\citeyear{besag1975}), where the joint probability of an outcome is replaced by a product of variable-wise conditional distributions:
\begin{equation} \label{eq:pseudo_likelihood}
    pl(x_V) = \prod\limits_{j = 1}^d p(x_j \mid x_{V \setminus j}).
\end{equation}
Under certain assumptions which generally hold if we assume that the data was generated from a Markov network, the pseudo-likelihood is a consistent estimator of the model parameters (Koller and Friedman \citeyear{koller2009}, Section 20.6.1). The major advantage of this approximation is that the full conditional distributions for each variable have a surprisingly simple form. Let $j \in V$ and consider the set of edges $E_j \subseteq E$ involving $j$. Then we have that
\begin{equation} \label{eq:full_conditionals}
    p(x_j \mid x_{V \setminus j}) = \frac{\exp\left\{\theta_j x_j + \sum_{(j, j^\prime) \in E_j} \theta_{j j^\prime} x_j x_{j^\prime}\right\}}{\sum_{x_j} \exp\left\{\theta_j x_j + \sum_{(j, j^\prime) \in E_j} \theta_{j j^\prime} x_j x_{j^\prime}\right\}} = p(x_j \mid x_{mb(j)}).
\end{equation}
Here the key observation is that the problematic global normalizing constant $Z$ disappears, and is in a sense replaced with local normalizing constants. Since Equation \eqref{eq:full_conditionals} only involves parameters associated with $X_j$, demonstrating the Local Markov Property \ref{itm:L}, it allows efficient computing of all the terms.

Moreover, Equation \eqref{eq:full_conditionals} is in the form of a logistic regression model. In the following, let $\theta_j x_j = \beta_{j 0} x_j$ and $\theta_{j j^\prime} x_j x_{j^\prime} = \beta_{j j^\prime} x_j x_{j^\prime}$. For observations $x_{1j}, \ldots, x_{Nj}$ and associated observations $x_{1j^\prime}, \ldots, x_{Nj^\prime}$ for all $j^\prime \in mb(j)$, the log-likelihood of the logistic regression model is
\begin{equation} \label{eq:lr_binomial}
    \log L(\beta_{j \mid mb(j)}) = \sum\limits_{i = 1}^N \Bigl(x_{ij} \log(\mu_{ij}) + (1 - x_{ij}) \log(1 - \mu_{ij})\Bigr),
\end{equation}
where
\begin{equation} \label{eq:mu}
    \mu_{ij} = p(x_{ij} = 1 \mid x_{i, mb(j)}, \beta_{j \mid mb(j)}) = \frac{\exp\Bigl\{\beta_{j 0} + \sum_{j^\prime \in mb(j)} \beta_{j j^\prime} x_{ij^\prime}\Bigr\}}{1 + \exp\Bigl\{\beta_{j 0} + \sum_{j^\prime\in mb(j)} \beta_{j j^\prime} x_{ij^\prime}\Bigr\}}.
\end{equation}
To treat the $d$ regression problems separately, we uncouple the parameter pairs by allowing $\beta_{j j^\prime} \neq \beta_{j^\prime j}$ as is done in Barber and Drton (\citeyear{barber2015}).

Since adding a variable to the Markov blanket (nearly) always increases the value of the maximum likelihood, it is necessary to incorporate a regularization term penalizing the complexity of the model. In particular, $L_1$-regularized logistic regression has been a popular choice, used for example by Ravikumar, Wainwright and Lafferty (\citeyear{ravikumar2010}). However, in this work we consider the extended Bayesian Information Criterion, which was recently proposed in the context of structure learning by Barber and Drton (\citeyear{barber2015}):
\begin{equation} \label{eq:bic}
    \text{BIC}_\gamma(\hat{\beta}_{j \mid mb(j)}) = \log L(\hat{\beta}_{j \mid mb(j)}) - \text{dim}(\hat{\beta}_{j \mid mb(j)}) \left(\frac{\log(N)}{2} + \gamma \log(d - 1)\right),
\end{equation}
where $\hat{\beta}_{j \mid mb(j)}$ is the maximum likelihood estimate (solved numerically). Here, BIC$_0$ is the well-known classical Bayesian information criterion (BIC) (Schwarz \citeyear{schwarz1978}). As discussed in \.{Z}ak-Szatkowska and Bogdan (\citeyear{zak-szatkowska2011}), the classical BIC has a tendency to choose too many parameters in the solution: namely, the asymptotic assumptions rely on $d$ being constant, which is inappropriate when $d$ is comparable to $N$ or larger. To remedy this, the extension term $\gamma \log(d - 1)$ was introduced as a prior on the set of considered models, where the choice of $\gamma \geq 0$ controls the strength of the prior. Based on the results of Barber and Drton (\citeyear{barber2015}), we fix $\gamma = 0.5$ in our numerical experiments.

Barber and Drton (\citeyear{barber2015}) proved that BIC$_\gamma$ is a consistent scoring function under certain sparsity conditions, which require that the maximum node degree $q$ grows sublinearly in $d$, that is
\begin{equation} \label{eq:sparsity}
    \vert mb(j) \vert \leq q = o(d), \quad \text{for all } j \in V.
\end{equation}
This assumption is common in the design of contemporary methods and can be justified by practical applications of graphical models where the condition is reasonable, e.g. image analysis and social networks (Ravikumar, Wainwright and Lafferty \citeyear{ravikumar2010}).

\section{Search algorithm} \label{section:search_algorithm}

\subsection{Search algorithms for pseudo-likelihood-based methods} \label{subsection:search_algorithm_background}
The task of estimating a graph structure $\hat{G} = (V, \hat{E})$ from data with respect to a scoring function is commonly split into two phases in the pseudo-likelihood framework. The first phase involves maximizing \eqref{eq:bic} locally for each variable $j \in V$. Then, since the collection of Markov blankets estimated in the first phase is in general not consistent with an undirected graph, a second phase is used to combine the $d$ local solutions into a globally consistent graph structure. The aim of this section is to establish notation for distinguishing between different optimization strategies. For each introduced method, the left-hand side of the name indicates the search algorithm and the right-hand side the associated scoring function. We begin by introducing two recent strategies. For the sake of clarity, the scoring function in each of them is exchangeably replaced with BIC$_\gamma$.
\begin{description}
    \item{\textbf{HC}$_{\lor / \land}$\textbf{-BIC}$_\gamma$} \\ \textbf{First phase.} Learn the Markov blanket $\widehat{mb}(j)$ for each variable $j \in V$ using a greedy hill climbing procedure similar to the IAMB algorithm (Tsamardinos et al. \citeyear{tsamardinos2003}). The hill climbing algorithm starts with $\widehat{mb}(j) = \emptyset$ and carries out addition and deletion iterations on the remaining $d - 1$ variables until no addition or deletion of a variable to the Markov blanket improves the score. This kind of an approach was used for example by Jalali, Johnson and Ravikumar (\citeyear{jalali2011}). \\ \textbf{Second phase.} Form the final estimated graph from the $d$ Markov blanket solutions by applying either an \textsc{OR} ($\lor$) or an \textsc{AND} ($\land$) rule (described below).
    \item{\textbf{L1LR}$_{\lor / \land}$\textbf{-BIC}$_\gamma$} \\ \textbf{First phase.} Produce a list of candidate Markov blankets for each variable $j \in V$ by running a series of $(d - 1)$-dimensional $L_1$-regularized logistic regressions with varying levels of penalization and collecting variables corresponding to nonzero $\beta$ parameter estimates (Ravikumar, Wainwright and Lafferty \citeyear{ravikumar2010}). Barber and Drton (\citeyear{barber2015}) then apply BIC$_\gamma$ to select the best Markov blanket from the candidate Markov blanket list. \\ \textbf{Second phase.} As above, the final estimated graph is formed using either the \textsc{OR} or \textsc{AND} rule.
\end{description}
With the \textsc{OR} rule, $\hat{E}_\lor = \{(j, j^\prime) \in \hat{E} : j \in \widehat{mb}(j^\prime) \lor j^\prime \in \widehat{mb}(j)\}$ and with the \textsc{AND} rule, $\hat{E}_\land = \{(j, j^\prime) \in \hat{E} : j \in \widehat{mb}(j^\prime) \land j^\prime \in \widehat{mb}(j)\}$ in the final estimated graph. 

To remove the $\lor / \land$ user choice, Pensar et al. (\citeyear{pensar2017}) introduced an alternative approach that makes further use of the underlying pseudo-likelihood score:
\begin{description}
    \item{\textbf{HC-BIC}$_\gamma$} \\ \textbf{First phase.} Obtain $\hat{E}_\lor$ by running HC$_\lor$-BIC$_\gamma$. \\ \textbf{Second phase.} Considering the first phase solution as a prescan that identifies eligible edges, apply another hill climbing procedure on $\hat{E}_\lor$ that in each iteration chooses the highest scoring neighboring graph structure (differing by 1 edge). Once the score can no longer be improved by a single edge change, return the final estimated graph.
\end{description}
Because of the variable-wise factorization, local edge changes in the second phase of HC-BIC$_\gamma$ cause a recalculation of the score for only two variables, meaning that each iteration can be carried out efficiently by caching the edge-wise score differences.

\subsection{Search algorithm using correlation neighborhoods} \label{subsection:search_algorithm_correlation_neighborhoods}
Our main contribution is a modification to the first phase of HC-BIC$_\gamma$. The particular issue we are going to address is that the IAMB algorithm run for any variable $j \in V$ iterates over the set $V \setminus \left(j \, \cup \, mb(j)\right)$ in each addition step and therefore, under the sparsity condition \eqref{eq:sparsity}, examines a considerable number of extra variables in each iteration. To counteract this, we propose forming small candidate neighborhoods, termed here correlation neighborhoods, for each variable with penalized likelihood ratio (PLR) tests by utilizing the correlation decay property (Bresler, Mossel and Sly \citeyear{bresler2008}). Bresler, Mossel and Sly (\citeyear{bresler2008}) show that, for sparse Ising models, the true neighborhoods are subsets of the formed correlation neighborhoods with high probability given a large enough threshold. In our approach the likelihood ratio can be thought of as a correlation measure for which the BIC$_\gamma$ penalty functions as a significance threshold. The full form of this strategy is as follows:
\begin{description}
    \item{\textbf{PLRHC-BIC}$_\gamma$}
    \\ \textbf{First phase.} Form correlation neighborhoods for each variable by running penalized likelihood ratio (PLR) tests for all variable pairs. That is, form the set 
        \begin{equation} \label{eq:pairwise_test}
            E^{PLR} = \{(j, j^\prime) \in V \times V : j \neq j^\prime, \text{BIC}_\gamma(\hat{\beta}_{j \mid mb(j) = j^\prime}) - \text{BIC}_\gamma(\hat{\beta}_{j \mid mb(j)=\emptyset}) > 0 \}.
        \end{equation}
Next, run HC$_\lor$-BIC$_\gamma$ with constrained search spaces collected from $E^{PLR}$ to obtain $\hat{E}_\lor$. 
    \\ \textbf{Second phase.} Apply the second phase of HC-BIC$_\gamma$ on $\hat{E}_\lor$ normally.
\end{description}
Additionally, we will further utilize the exponential decay of correlations as a function of graph-theoretic distance by constructing the constrained search space of each variable $j$ so that it contains all nodes within a distance of 3 from $j$ with respect to $E^{PLR}$. The motivation for this is that if the tests for $j$ did not pass for all true neighbors, the corresponding tests of $j$'s neighbors (or their neighbors) may have passed for other true neighbors of $j$. This way, we increase the probability of capturing all the true neighbors of $j$ while still appropriately limiting the size of the correlation neighborhoods.

We note that in the asymptotic scenario where $N$ tends to infinity and $d$ is fixed, the threshold given by the BIC$_\gamma$ penalty will ultimately add all possible edges within each (true) graph component to $E^{PLR}$ since blocking a node from the rest of the network requires conditioning on its (true) Markov blanket. However, this will not happen in practice for realistic sample sizes.

\section{Numerical experiments} \label{section:experiments}
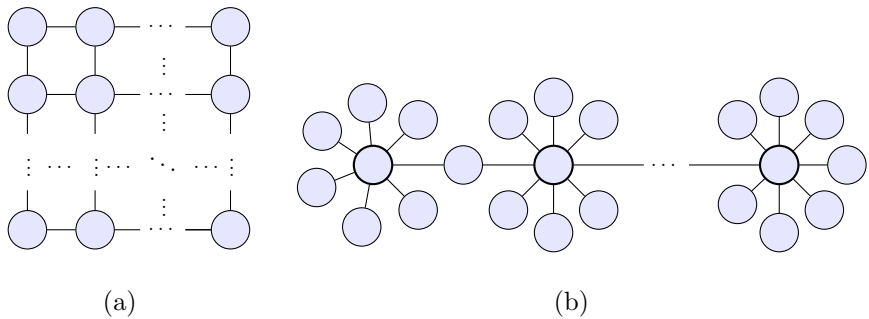
\begin{figure}
\captionsetup{font=scriptsize}
\hspace*{-1cm}
\centering 
\subfigure[]{\label{fig:grid}
\begin{tikzpicture}[scale=.3, auto=left, every node/.style={circle, fill=blue!10, draw=black, scale=.7}]
    \node(n1) at (0, 9) {\phantom{0}};
    \node(n2) at (3, 9) {\phantom{0}};
    \coordinate(n31) at (5, 9);
    \coordinate(n32) at (7, 9);
    \node(n4) at (9, 9) {\phantom{0}};
    \node(n5) at (0, 6) {\phantom{0}};
    \node(n6) at (3, 6) {\phantom{0}};
    \coordinate(n71) at (5, 6);
    \coordinate(n72) at (7, 6);
    \node(n8) at (9, 6) {\phantom{0}};
    \node(n13) at (0, 0) {\phantom{0}};
    \node(n14) at (3, 0) {\phantom{0}};
    \coordinate(n111) at (5, 3);
    \coordinate(n112) at (7, 3);
    \coordinate(n141) at (5, 0);
    \coordinate(n142) at (7, 0);
    \coordinate(n5down) at (0, 4.25);
    \coordinate(n6down) at (3, 4.25);
    \coordinate(n8down) at (9, 4.25);
    \coordinate(n13up) at (0, 1.75);
    \coordinate(n14up) at (3, 1.75);
    \coordinate(n16up) at (9, 1.75);
    \coordinate(n16left) at (7, 0);
    \node(n16) at (9, 0) {\phantom{0}};
    \path (5, 8) -- node[draw=none, fill=none]{\ldots} (7, 8);
    \path (5, 5) -- node[draw=none, fill=none]{\ldots} (7, 5);
    \path (5, -1) -- node[draw=none, fill=none]{\ldots} (7, -1);
    \path (0.5, 1.75) -- node[draw=none, fill=none]{\ldots} (2.5, 1.75);
    \path (3.25, 1.75) -- node[draw=none, fill=none]{\ldots} (5, 1.75);
    \path (7, 1.75) -- node[draw=none, fill=none]{\ldots} (9, 1.75);
    \path (5, 1.75) -- node[draw=none, fill=none]{\large{$\ddots$}} (7, 1.75);
    \path (5, 9) -- node[draw=none, fill=none]{\vdots} (5, 6);
    \path (5, 6) -- node[draw=none, fill=none]{\vdots} (5, 4);
    \path (5, 2.5) -- node[draw=none, fill=none]{\vdots} (5, 0);
    \path (-1, 4) -- node[draw=none, fill=none]{\vdots} (-1, 2);
    \path (2, 4) -- node[draw=none, fill=none]{\vdots} (2, 2);
    \path (8, 4) -- node[draw=none, fill=none]{\vdots} (8, 2);
    \foreach \from/\to in {n1/n2, n1/n5, n2/n6, n8/n4, n14/n13, n6/n5}
    \draw (\from) -- (\to);
    \foreach \from/\to in {n2/n31, n32/n4, n6/n71, n72/n8, n14/n141, n142/n16}
    \draw (\from) -- (\to);
    \foreach \from/\to in {n5/n5down, n6/n6down, n8/n8down, n13up/n13, n14up/n14, n16up/n16, n16left/n16}
    \draw (\from) -- (\to);
\end{tikzpicture}}\quad
\subfigure[]{\label{fig:hub}
\begin{tikzpicture}[scale=.3, auto=left, every node/.style={circle, fill=blue!10, draw=black, scale=.7}]
    \node[thick](n1) at (21, 3) {\phantom{0}};
    \node(n2) at (19.15, 1.25) {\phantom{0}};
    \node(n3) at (21, 0) {\phantom{0}};
    \node(n4) at (23, 1) {\phantom{0}};
    \node(n5) at (24, 3) {\phantom{0}};
    \node(n6) at (23, 5) {\phantom{0}};
    \node(n7) at (21, 6) {\phantom{0}};
    \node(n8) at (19.15, 5) {\phantom{0}};
    \node(share1) at (7, 3) {\phantom{0}};
    \node[thick](mid) at (11, 3) {\phantom{0}};
    \node(midn1) at (13, 5) {\phantom{0}};
    \node(midn2) at (11, 6) {\phantom{0}};
    \node(midn3) at (9, 5) {\phantom{0}};
    \node(midn4) at (9, 1) {\phantom{0}};
    \node(midn5) at (11, 0) {\phantom{0}};
    \node(midn6) at (13, 1) {\phantom{0}};
    \node[draw=none, fill=none](share2) at (15, 3) {\phantom{0}};
    \coordinate(share2c1) at (15, 3);
    \path (17, 3) -- node[draw=none, fill=none]{\ldots} (17, 3);
    \coordinate(share2c2) at (17, 3);
    \node[thick](n9) at (3, 3) {\phantom{0}};
    \node(n10) at (5, 5) {\phantom{0}};
    \node(n11) at (2.75, 5.75) {\phantom{0}};
    \node(n12) at (0.75, 4.5) {\phantom{0}};
    \node(n13) at (0.5, 2) {\phantom{0}};
    \node(n14) at (2.5, 0.25) {\phantom{0}};
    \node(n15) at (5, 1) {\phantom{0}};
    \foreach \from/\to in {n1/n2, n1/n3, n1/n4, n1/n5, n1/n6, n1/n7, n1/n8, n9/share1, share1/mid, mid/share2c1, share2c2/n1,
                mid/midn1, mid/midn2, mid/midn3, mid/midn4, mid/midn5, mid/midn6, n9/n10, n9/n11, n9/n12, n9/n13, n9/n14, n9/n15}
    \draw (\from) -- (\to);
\end{tikzpicture}}
\caption{Synthetic Markov network structures. (a) Grid network. (b) Hub network with hub nodes (drawn with thick border for emphasis).}
\label{fig:synthetic_networks}
\end{figure}

In this section, we study the performance of PLRHC-BIC$_{0.5}$ on binary data generated from synthetic Markov networks with known true structure. We consider the two connected commonly-used graph types shown in Figure~\ref{fig:synthetic_networks}. The grid network is a common benchmark and the hub network is considered difficult to learn because of the high-degree hub nodes (Ravikumar, Wainwright and Lafferty \citeyear{ravikumar2010}; Pensar et al. \citeyear{pensar2017}). We generate the distribution of a synthetic Markov network by sampling edge potential values independently from a standard uniform distribution. Next, we simulate a data set from this distribution using Gibbs sampling with a burn-in set to 100\,000 and thinning to 100. For each experiment in this section we generated 100 data sets per desired sample size $N$ and averaged the results.

With the true structure of the Markov network known, we measure the quality of a solution by the number of mislearned edges as follows. Let $G$ be the true graph with edges $E$ and $\hat{G}$ the learned graph with edges $\hat{E}$. The number of false positive edges is given by $FP(G, \hat{G}) = \vert \hat{E} \setminus E \vert$, the number of false negative edges by $FN(G, \hat{G}) = \vert E \setminus \hat{E} \vert$ and the Hamming distance between the learned and true graph by $HD(G, \hat{G}) = FP(G, \hat{G}) + FN(G, \hat{G})$. In practice, the number of edges varies drastically according to graph type and number of nodes $d$, so we consider also a standardized Hamming distance, which we define as
\begin{equation} \label{eq:std_hamming_distance}
    HD_{\text{std}}(G, \hat{G}) = 100 \cdot \frac{HD(G, \hat{G})}{\vert E \vert}.
\end{equation}
This gives us a normalized measure of solution quality independent of the size of the graph or the number of edges in the graph.

\subsection{Reliability of penalized likelihood ratio tests}
\begin{table}
\captionsetup{font=scriptsize}
\centering
{\scriptsize
\begin{tabular}{@{}r@{\hskip3pt}|c@{\hskip7pt}cccccc@{\hskip7pt}l}
\multicolumn{9}{c}{Grid network ($d = 256$)} \\
\cline{1-8}
Sample size ($N$) & $250$ & $500$ & $1\,000$ & $2\,000$ & $4\,000$ & $8\,000$ & $16\,000$ & \\
\cline{1-8}
HC$_\lor$-BIC$_{0.5}$ recall & 39.0 \% & 50.7 \% & 60.6 \% & 69.7 \% & 77.1 \% & 82.6 \% & 87.3 \% & \\
$\vert E^{PLR}_1 \vert \, / \, \vert D_1 \vert$ & 40.4 \% & 52.2 \% & 62.1 \% & 71.0 \% & 78.3 \% & 83.4 \% & 88.2 \% & \, $\vert D_1 \vert = 960$ \\
\cline{1-8}
$\vert E^{PLR}_2 \vert \, / \, \vert D_2 \vert$ & 6.93 \% & 11.7 \% & 17.6 \% & 25.8 \% & 33.5 \% & 41.9 \% & 50.9 \% & \, $\vert D_2 \vert = 1\,796$ \\
$\vert E^{PLR}_4 \vert \, / \, \vert D_3 \vert$ & 0.79 \% & 1.55 \% & 2.78 \% & 5.05 \% & 7.92 \% & 12.0 \% & 17.5 \% & \, $\vert D_3 \vert = 2\,512$ \\
$\vert E^{PLR}_3 \vert \, / \, \vert D_4 \vert$ & 0.15 \% & 0.19 \% & 0.35 \% & 0.68 \% & 1.16 \% & 2.13 \% & 3.95 \% & \, $\vert D_4 \vert = 3\,112$ \\
$\vert E^{PLR}_{\geq 5} \vert \, / \, \vert D_{\geq 5} \vert$ & 0.09 \% & 0.07 \% & 0.05 \% & 0.04 \% & 0.03 \% & 0.03 \% & 0.06 \% & \, $\vert D_{\geq 5} \vert = 56\,900$ \\
\cline{1-8}
$\vert E^{PLR} \vert$ & 588 & 793 & 1\,020 & 1\,314 & 1\,604 & 1\,939 & 2\,362 & \, $\vert D_{\geq 1} \vert =65\,280$ \\
\multicolumn{9}{l}{} \\
\multicolumn{9}{c}{Hub network ($d = 512$)} \\
\cline{1-8}
Sample size ($N$) & $250$ & $500$ & $1\,000$ & $2\,000$ & $4\,000$ & $8\,000$ & $16\,000$ & \\
\cline{1-8}
HC$_\lor$-BIC$_{0.5}$ recall & 40.2 \% & 51.8 \% & 61.6 \% & 70.8 \% & 77.7 \% & 83.7 \% & 88.4 \% & \\
$\vert E^{PLR}_1 \vert \, / \, \vert D_1 \vert$ & 40.5 \% & 52.1 \% & 61.9 \% & 71.0 \% & 77.9 \% & 83.8 \% & 88.5 \% & \, $\vert D_1 \vert = 1\,022$ \\
\cline{1-8}
$\vert E^{PLR}_2 \vert \, / \, \vert D_2 \vert$ & 5.20 \% & 9.07 \% & 14.1 \% & 20.4 \% & 27.6 \% & 35.2 \% & 43.7 \% & \, $\vert D_2 \vert = 3\,696$ \\
$\vert E^{PLR}_3 \vert \, / \, \vert D_3 \vert$ & 0.34 \% & 0.53 \% & 1.02 \% & 2.13 \% & 3.57 \% & 5.93 \% & 9.19 \% & \, $\vert D_3 \vert = 1\,762$ \\
$\vert E^{PLR}_4 \vert \, / \, \vert D_4 \vert$ & 0.09 \% & 0.06 \% & 0.08 \% & 0.19 \% & 0.32 \% & 0.65 \% & 1.24 \% & \, $\vert D_4 \vert = 6\,284$ \\
$\vert E^{PLR}_{\geq 5} \vert \, / \, \vert D_{\geq 5} \vert$ & 0.07 \% & 0.04 \% & 0.03 \% & 0.02 \% & 0.01 \% & 0.01 \% & 0.01 \% & \, $\vert D_{\geq 5} \vert = 248\,868$ \\
\cline{1-8}
$\vert E^{PLR} \vert$ & 781 & 988 & 1\,250 & 1\,581 & 1\,930 & 2\,327 & 2\,776 & \, $\vert D_{\geq 1} \vert = 261\,632$ \\
\end{tabular}
}
\caption{The inclusion rate of true edges in $E^{PLR}$ against the recall of HC$_\lor$-BIC$_{0.5}$ and size of the reduced edge set $E^{PLR}$. The inclusion rate of edges in $E^{PLR}$ is grouped based on the graph-theoretic distance between the node pairs in the true graph $G$. Let $D_k$ contain the node pairs whose shortest distance between them in $G$ is $k$ and $D_{\geq k}$ the pairs whose shortest distance is $k$ or more. Let $E^{PLR}_k$ be the node pairs that are in both $E^{PLR}$ and $D_k$ and similarly for $E^{PLR}_{\geq k}$. Then $\vert E^{PLR}_k \vert \, / \, \vert D_k \vert$ is the inclusion rate of node pairs into $E^{PLR}$ that are $k$ distance apart in $G$ and $\vert E^{PLR}_1 \vert \, / \, \vert D_1 \vert$ is the inclusion rate of true edges. The results are averaged over 100 sampled data sets.}
\label{table:netest}
\end{table}

In the first experiment, our goal was to study the reliability and size of the reduced edge set $E^{PLR}$ (see Section~\ref{subsection:search_algorithm_correlation_neighborhoods}) formed with the penalized likelihood ratio (PLR) tests. To measure reliability, we used the recall, that is the proportion of true edges that were successfully retrieved, of HC$_\lor$-BIC as a baseline and compared the inclusion rate of true edges in $E^{PLR}$ against this baseline. The size, $\vert E^{PLR} \vert$, was compared against the size of the set of all possible pairs to see whether the reduction in size would be considerable. We also examined the node pairs included in $E^{PLR}$ with regard to their graph-theoretic distance in the true graph. The experiments were run on data sets generated from a grid network ($d = 256$) and a hub network ($d = 512$) with sample sizes ranging from $N = 250$ to $N = 16\,000$ in both cases. The results are collected in Table~\ref{table:netest}.

Looking at the results, we see that approximately the same proportion of true edges were included in $E^{PLR}$ when compared with the baseline. In fact, the compared proportions were slightly larger for $E^{PLR}$ in all test cases. This likely happened because in the full run of the HC$_\lor$-BIC$_{0.5}$ algorithm the penalty term regularizes the complexity of the model when the Markov blankets grow in size. This results in a loss of a few of the weaker true edges but in return greatly reduces the number of false positive edges. We also observed the exponential decay as a function of distance with fewer and fewer node pairs of increasing distance in the true graph being included in $E^{PLR}$. The inclusion rate was virtually 0 \% for node pairs that are distant in the true graph, which means that the PLR tests successfully obtained a considerably smaller, but accurate, reduced edge set $E^{PLR}$ than the set of all possible pairs.

\subsection{Computational cost and accuracy}
\begin{table}
\captionsetup{font=scriptsize}
\centering
{\scriptsize
\begin{tabular}{lr|ccccc}
\multicolumn{7}{c}{Grid network ($N = 4\,000$)} \\
\hline
\multicolumn{2}{r|}{Number of variables ($d$)} & $64$ & $144$ & $256$ & $400$ & $1\,024$ \\
\hline
HC-BIC$_{0.5}$: & Number of iterations ($\times 1\,000$) & 15.2 & 78.9 & 252 & 622 & 4\,142 \\
\multicolumn{2}{r|}{Proportion of iterations involving only two variables} & 26.8 \% & 26.2 \% & 26 \% & 25.7 \% & 25.3 \% \\
\multicolumn{2}{r|}{Average Markov blanket size per iteration} & 2.44 & 2.5 & 2.53 & 2.55 & 2.59 \\
\hline
PLRHC-BIC$_{0.5}$: & Number of iterations ($\times 1\,000$) & 11 & 40.6 & 105 & 226 & 1\,245 \\
\multicolumn{2}{r|}{Proportion of iterations involving only two nodes} & 51.9 \% & 62.5 \% & 71.3 \% & 77.8 \% & 88.3 \% \\
\multicolumn{2}{r|}{Average Markov blanket size per iteration} & 1.93 & 1.75 & 1.58 & 1.45 & 1.24 \\
\hline
\multicolumn{2}{r|}{$HD_{std}(G, \hat{G}_{PLRHC-BIC_{0.5}}) - HD_{std}(G, \hat{G}_{HC-BIC_{0.5}})$} & 0.01 & 0.19 & 0.3 & 0.32 & 0.44 \\
\multicolumn{7}{l}{} \\
\multicolumn{7}{c}{Hub network ($N = 4\,000$)} \\
\hline
\multicolumn{2}{r|}{Number of variables ($d$)} & $64$ & $128$ & $256$ & $400$ & $1\,024$ \\
\hline
HC-BIC$_{0.5}$: & Number of iterations to obtain $\hat{E}_\lor$ ($\times 1\,000$) & 10.4 & 52.4 & 166 & 407 & 2\,694 \\
\multicolumn{2}{r|}{Proportion of iterations involving only two variables} & 39.3 \% & 39.5 \% & 39.4 \% & 39.3 \% & 38.9 \% \\
\multicolumn{2}{r|}{Average Markov blanket size per iteration} & 2.34 & 2.35 & 2.36 & 2.37 & 2.38 \\
\hline
PLRHC-BIC$_{0.5}$: & Number of iterations to obtain $\hat{E}_\lor$ ($\times 1\,000$) & 7.2 & 29.7 & 86 & 202 & 1\,260 \\
\multicolumn{2}{r|}{Proportion of iterations involving only two variables} & 74.9 \% & 86.1 \% & 91.3 \% & 94 \% & 97.4 \% \\
\multicolumn{2}{r|}{Average Markov blanket size per iteration} & 1.51 & 1.29 & 1.18 & 1.12 & 1.05 \\
\hline
\multicolumn{2}{r|}{$HD_{std}(G, \hat{G}_{PLRHC-BIC_{0.5}}) - HD_{std}(G, \hat{G}_{HC-BIC_{0.5}})$} & 0.05 & 0.15 & 0.22 & 0.33 & 0.93 \\
\end{tabular}
}
\caption{Computational cost and accuracy comparison between PLRHC-BIC$_{0.5}$ and HC-BIC$_{0.5}$ when learning grid and hub networks of varying sizes. Results are averaged over 100 sampled data sets.}
\label{table:cost}
\end{table}

In the second experiment, we compared the computational cost and accuracy of PLRHC-BIC$_{0.5}$ and HC-BIC$_{0.5}$. The experiments were run on data sets generated from grid and hub networks with the number of variables ranging from $d = 64$ to $d = 1\,024$ and the sample size fixed to $N = 4\,000$. The computational cost was measured in the number of hill-climbing iterations performed by each algorithm and the proportion of iterations involving only two variables (that is, iterations with Markov blankets of a single variable). The proportion of pairwise iterations is of interest because the maximum likelihood estimator $\hat{\beta}_{j \mid mb(j)}$ is numerically solved faster for small Markov blankets. We also report the average Markov blanket size per iteration. The accuracy of the estimated graph structures $\hat{G}_{PLRHC-BIC_{0.5}}$ and $\hat{G}_{HC-BIC_{0.5}}$ was measured by their standardized Hamming distance to the true graph structure $G$. The results are collected in Table~\ref{table:cost}.

Looking at the results of PLRHC-BIC$_{0.5}$, the constrained search spaces obtained with the penalized likelihood ratio (PLR) tests reduced considerably the number of iterations required to estimate $\hat{G}_{PLRHC-BIC_{0.5}}$ when compared with the number of iterations HC-BIC$_{0.5}$ took to estimate $\hat{G}_{HC-BIC_{0.5}}$. Additionally, most of the iterations involved only two variables for PLRHC-BIC$_{0.5}$. We also observed a slight improvement in learning accuracy with PLRHC-BIC$_{0.5}$, which comes as a reduction of false positive edges in the estimated graph structure. This is likely due to the PLR tests eliminating noisy edges from the search spaces which could cause the IAMB algorithm to overfit. As $d$ was increased, the gain in both computational efficiency and accuracy became more pronounced.

\subsection{Comparative evaluation}
\begin{figure}
\captionsetup{font=scriptsize}
\centering
\includegraphics[width=7cm, height=7cm]{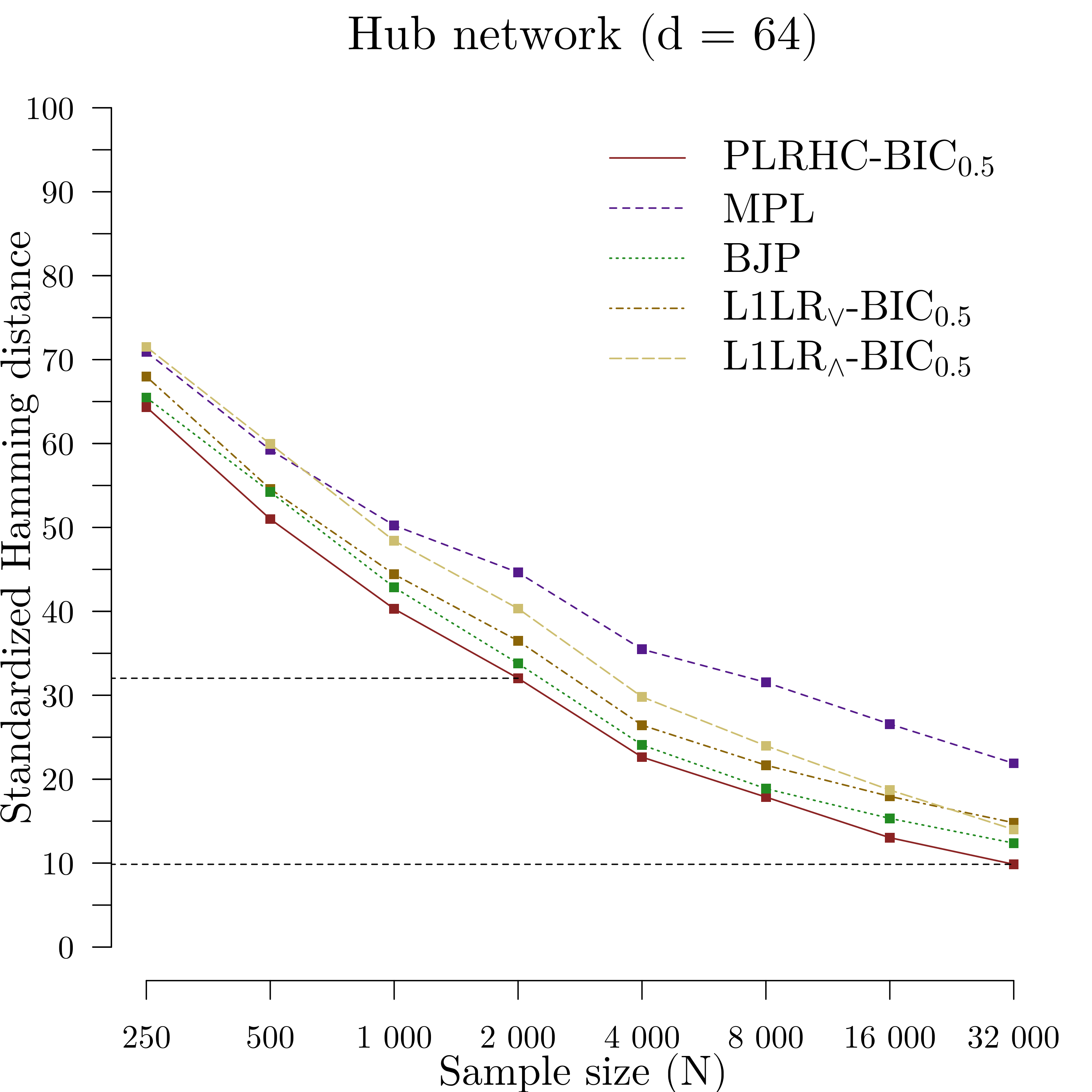}
\includegraphics[width=7cm, height=7cm]{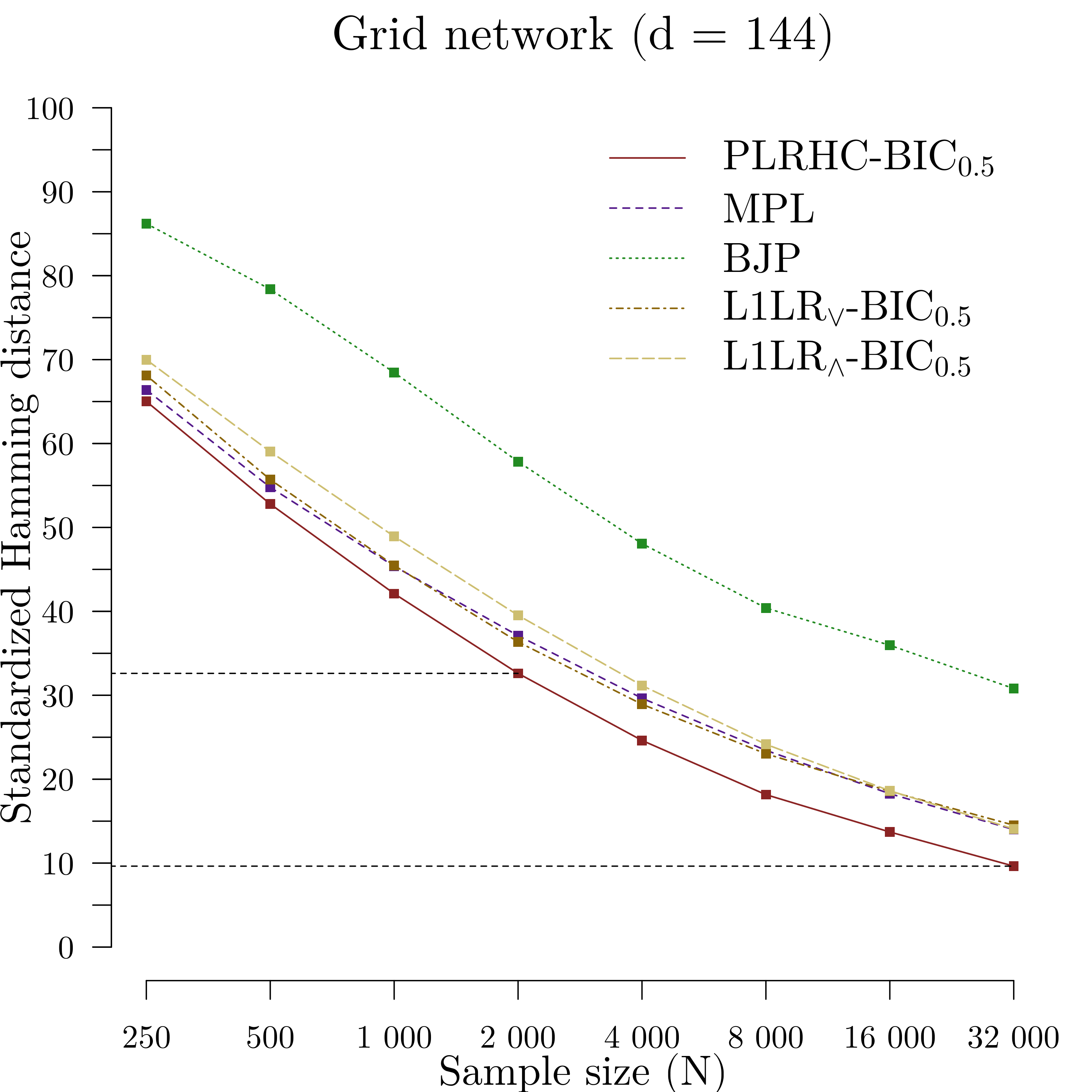}
\caption{Results for the smaller synthetic network experiments. Reference lines drawn horizontally for the smallest standardized Hamming distance at $N = 2\,000$ and $N = 32\,000$.}
\label{fig:comparison}
\end{figure}

\begin{figure}
\captionsetup{font=scriptsize}
\centering
\includegraphics[width=7cm, height=7cm]{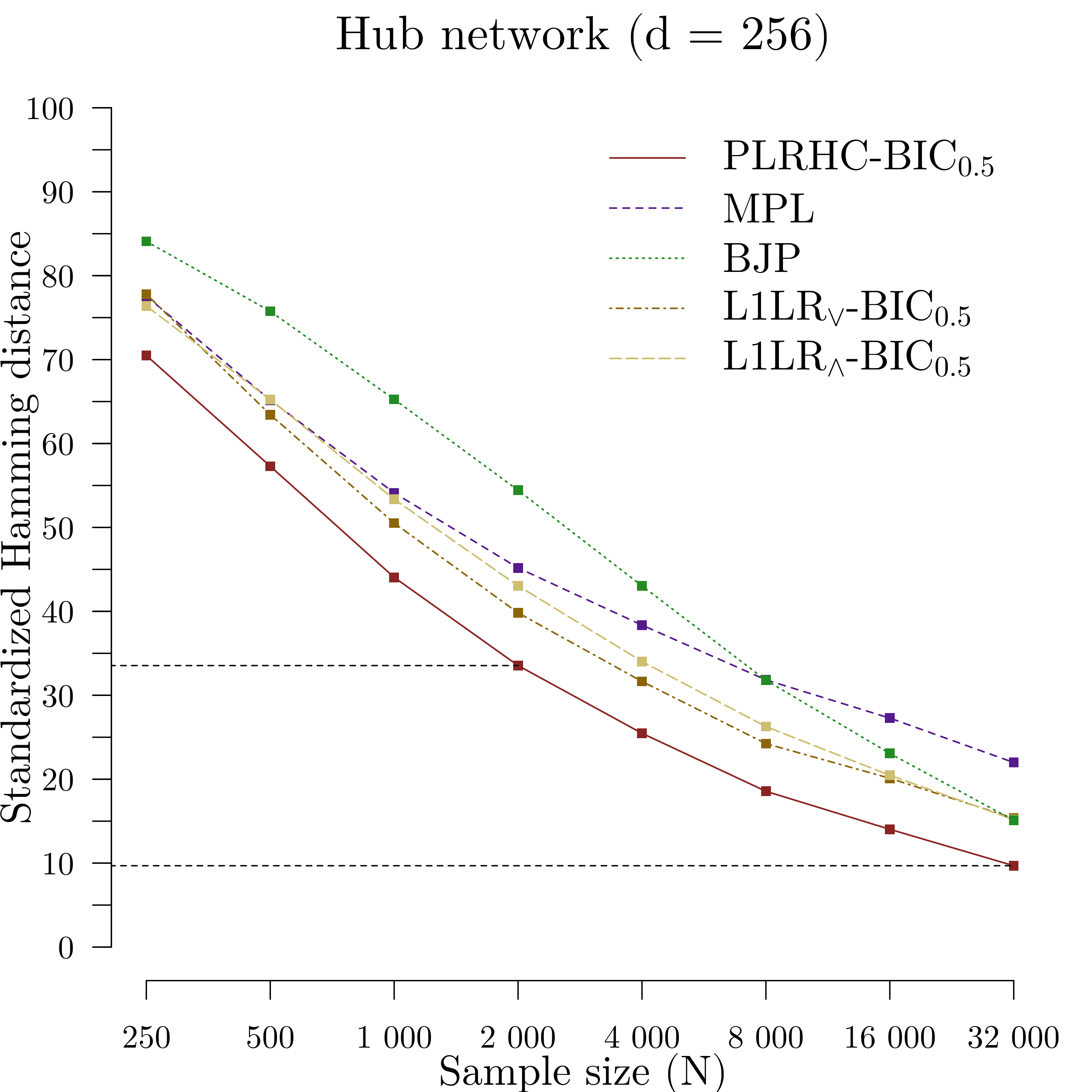}
\includegraphics[width=7cm, height=7cm]{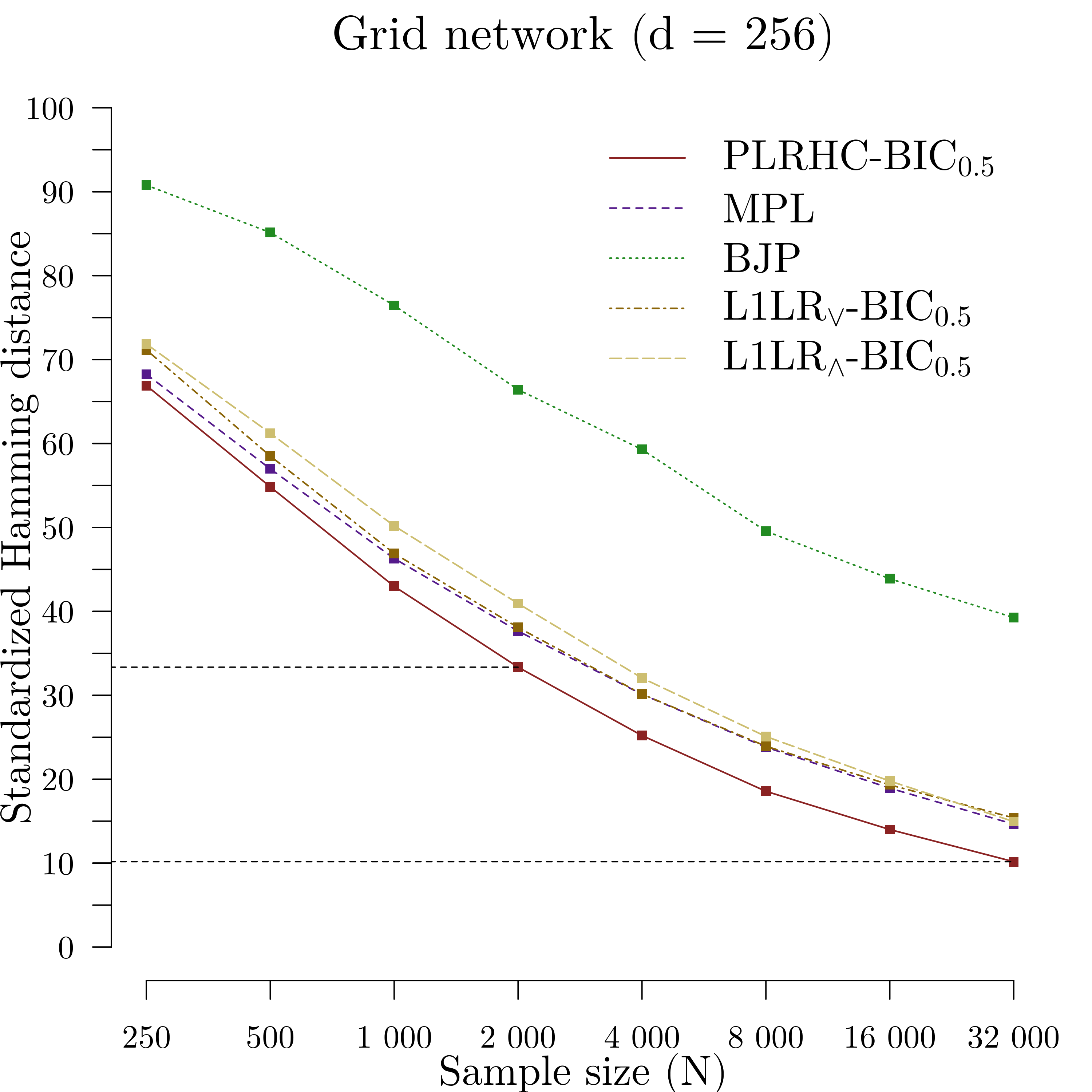}
\caption{Results for the larger synthetic network experiments. Reference lines drawn horizontally for the smallest standardized Hamming distance at $N = 2\,000$ and $N = 32\,000$.}
\label{fig:comparison2}
\end{figure}

In the final experiment, we compared the new search algorithm PLRHC-BIC$_{0.5}$ against L1LR$_{\lor / \land}$-BIC$_{0.5}$ by Barber and Drton (\citeyear{barber2015}). We additionally included two other recent Markov network structure learning methods in the comparison:
\begin{description}
    \item{\textbf{MPL.}} Pensar et al. (\citeyear{pensar2017}) apply the classical Bayesian-Dirichlet score (Heckerman, Geiger and Chickering \citeyear{heckerman1995}) for Markov networks by assuming a Multinomial-Dirichlet model for the pseudo-likelihood conditional distributions to get Marginal Pseudo-Likelihood (MPL). The structure prior in MPL score, which functions as an additional regularizer, is inspired by extended BIC. The MPL-optimal structure is learned with HC-MPL (see Section~\ref{subsection:search_algorithm_background}).
    \item{\textbf{BJP.}} Schl\"{u}ter et al. (\citeyear{schluter2018}) formulate the score of a graph as a joint probability distribution of Markov blankets. In the Blankets Joint Posterior (BJP) score, the posterior of each Markov blanket is computed with the chain rule progressively by using information from previously computed blankets as evidence. For high dimensions, the BJP score is approximated with the IBMAP-HC greedy hill climbing algorithm (Schl\"{u}ter, Bromberg and Edera \citeyear{schluter2014}).
\end{description}
The methods were compared with respect to the false positive and false negative edge counts, Hamming distance and the standardized Hamming distance of the estimated graph structures. The experiments were run on data sets generated from four networks: hub $(d = 64)$, grid $(d = 144)$, hub $(d = 256)$ and grid $(d = 256)$, and with sample sizes ranging from $N = 500$ to~$N = 32\,000$.

The standardized Hamming distances \eqref{eq:std_hamming_distance} of the estimated graph structures against the true structure are plotted in Figures~\ref{fig:comparison}~and~\ref{fig:comparison2} for each method. PLRHC-BIC$_{0.5}$
outperformed the other methods by achieving the highest accuracy, that is the smallest Hamming distance, throughout the experiments, no matter the type of network or sample size. Notably, the other methods suffered a larger loss in learning accuracy as a result of the increasing variable size $d$ for the two larger networks. In the larger hub network, BJP required a lot of data before its learning accuracy started approaching the other methods and the grid network appears to be an adversary network for BJP -- perhaps because, as the authors mention, BJP is designed to learn irregular graph structures while the grid network is very regular. As discussed by Pensar et al. (\citeyear{pensar2017}), the data efficiency issues emerging when learning hub networks with MPL were observable here for both hub networks.

Tables~\ref{table:comparison}~and~\ref{table:comparison2} contain a more detailed breakdown of the false positive (FP) and false negative (FN) edges for each method. PLRHC-BIC$_{0.5}$ obtained low FP values, competitive with the other methods, without losing too many true edges as indicated by the low FN values. In comparison, L1LR$_{\lor / \land}$-BIC$_{0.5}$ could achieve only one of the aforementioned properties: a low FN but high FP with the $\lor$ rule or a high FN but low FP with the $\land$ rule. MPL and PLRHC-BIC$_{0.5}$ performed similarly, but with MPL having higher FN values. BJP was careful at avoiding FP edges, but with larger $d$ it required a lot of data to reduce the FN value.

\begin{table}
\captionsetup{font=scriptsize}
\centering
{\scriptsize
\begin{tabular}{crccccccccc}
\multicolumn{10}{c}{Hub network $(d = 64)$} \\
\multicolumn{2}{c}{Method \textbackslash $\frac{N}{1000}$} & .25 & .5 & 1 & 2 & 4 & 8 & 16 & 32 \\
\toprule
\multirow{3}{*}{\begin{tabular}{@{}c} PLRHC- \\ BIC$_{0.5}$ \end{tabular}} & FP & 2.74 & 1.75 & 1.18 & 0.90 & 0.59 & 0.46 & 0.32 & 0.26 \\
                       & FN & 37.80 & 30.38 & 24.20 & 19.28 & 13.67 & 10.80 & 7.89 & 5.95 \\
                       & HD & 40.54 & 32.13 & 25.38 & 20.18 & 14.26 & 11.26 & 8.21 & 6.21 \\
\midrule
\multirow{3}{*}{\begin{tabular}{@{}c} L1LR$_\lor$ \\ -BIC$_{0.5}$ \end{tabular}} & FP & 6.92 & 5.29 & 4.54 & 4.53 & 3.58 & 3.29 & 3.64 & 3.48 \\
                       & FN & 35.91 & 29.10 & 23.44 & 18.46 & 13.08 & 10.36 & 7.67 & 5.85 \\
                       & HD & 42.83 & 34.39 & 27.98 & 22.99 & 16.66 & 13.65 & 11.31 & 9.33 \\
\midrule
\multirow{3}{*}{\begin{tabular}{@{}c} L1LR$_\land$ \\ -BIC$_{0.5}$ \end{tabular}} & FP & 1.64 & 0.96 & 0.65 & 0.60 & 0.43 & 0.30 & 0.18 & 0.20 \\
                       & FN & 43.41 & 36.81 & 29.84 & 24.81 & 18.35 & 14.81 & 11.61 & 8.62 \\
                       & HD & 45.05 & 37.77 & 30.49 & 25.41 & 18.78 & 15.11 & 11.79 & 8.82 \\
\midrule
\multirow{3}{*}{MPL} & FP & 4.38 & 2.38 & 1.25 & 0.91 & 0.35 & 0.28 & 0.15 & 0.08 \\
                       & FN & 40.30 & 34.94 & 30.40 & 27.21 & 22.00 & 19.59 & 16.59 & 13.71 \\
                       & HD & 44.68 & 37.32 & 31.65 & 28.12 & 22.35 & 19.87 & 16.74 & 13.79 \\
\midrule
\multirow{3}{*}{BJP} & FP & 0.78 & 0.63 & 0.60 & 0.86 & 0.66 & 1.16 & 1.81 & 1.94 \\
                       & FN & 40.47 & 33.54 & 26.40 & 20.43 & 14.51 & 10.73 & 7.85 & 5.86 \\
                       & HD & 41.25 & 34.17 & 27.00 & 21.29 & 15.17 & 11.89 & 9.66 & 7.80 \\
\multicolumn{10}{l}{} \\
\multicolumn{10}{c}{Grid network $(d = 144)$} \\
\multicolumn{2}{c}{Method \textbackslash $\frac{N}{1000}$} & .25 & .5 & 1 & 2 & 4 & 8 & 16 & 32 \\
\toprule
\multirow{3}{*}{\begin{tabular}{@{}c} PLRHC- \\ BIC$_{0.5}$ \end{tabular}} & FP & 6.17 & 4.12 & 2.49 & 1.58 & 1.18 & 0.78 & 0.67 & 0.31 \\
                       & FN & 165.47 & 135.25 & 108.69 &  84.50 &  63.82 &  47.17 &  35.53 &  25.14 \\
                       & HD & 171.64 & 139.37 & 111.18 &  86.08 &  65.00 &  47.95 &  36.20 &  25.45 \\
\midrule
\multirow{3}{*}{\begin{tabular}{@{}c} L1LR$_\lor$ \\ -BIC$_{0.5}$ \end{tabular}} & FP & 21.78 & 17.72 & 15.51 & 14.69 & 15.15 & 14.78 & 14.08 & 13.62 \\
                       & FN & 158.05 & 129.33 & 104.53 &  81.29 &  61.22 &  46.02 &  34.91 &  24.72 \\
                       & HD & 179.83 & 147.05 & 120.04 &  95.98 &  76.37 &  60.80 &  48.99 &  38.34 \\
\midrule
\multirow{3}{*}{\begin{tabular}{@{}c} L1LR$_\land$ \\ -BIC$_{0.5}$ \end{tabular}} & FP & 2.13 & 1.45 & 1.04 & 0.66 & 0.44 & 0.40 & 0.33 & 0.20 \\
                       & FN & 182.65 & 154.38 & 128.13 & 103.70 &  81.82 &  63.39 &  48.82 &  36.98 \\
                       & HD & 184.78 & 155.83 & 129.17 & 104.36 &  82.26 &  63.79 &  49.15 &  37.18 \\
\midrule
\multirow{3}{*}{MPL} & FP & 7.83 & 3.40 & 1.33 & 0.50 & 0.14 & 0.05 & 0.03 & 0.01 \\
                       & FN & 167.44 & 141.21 & 118.47 &  97.47 &  78.14 &  61.83 &  48.20 &  36.96 \\
                       & HD & 175.27 & 144.61 & 119.80 &  97.97 &  78.28 &  61.88 &  48.23 &  36.97 \\
\midrule
\multirow{3}{*}{BJP} & FP & 0.10 &  0.07 &  0.29 &  0.32 &  0.56 &  1.00 &  1.25 & 1.34 \\ 
                       & FN & 227.46 & 206.86 & 180.44 & 152.37 & 126.35 & 105.65 &  93.68 &  79.98 \\
                       & HD & 227.56 & 206.93 & 180.73 & 152.69 & 126.91 & 106.65 &  94.93 &  81.32 
\end{tabular}
}
\caption{Detailed results for the smaller synthetic network experiments. FP = number of false positive edges, FN = number of false negative edges and HD = Hamming distance. Results are averaged over 100 sampled data sets.}
\label{table:comparison}
\end{table}

\clearpage

\begin{table}
\captionsetup{font=scriptsize}
\centering
{\scriptsize
\begin{tabular}{crccccccccc}
\multicolumn{10}{c}{Grid network $(d = 256)$} \\
\multicolumn{2}{c}{Method \textbackslash $\frac{N}{1000}$} & .25 & .5 & 1 & 2 & 4 & 8 & 16 & 32 \\
\toprule
\multirow{3}{*}{\begin{tabular}{@{}c} PLRHC- \\ BIC$_{0.5}$ \end{tabular}} & FP & 12.34 & 7.77 & 5.10 & 3.60 & 2.79 & 1.44 & 1.09 & 0.87 \\ 
                       & FN & 308.84 & 255.45 & 201.22 & 156.47 & 118.16 & 87.66 & 66.11 & 47.95 \\ 
                       & HD & 321.18 & 263.22 & 206.32 & 160.07 & 120.95 & 89.10 & 67.20 & 48.82 \\
\midrule
\multirow{3}{*}{\begin{tabular}{@{}c} L1LR$_\lor$ \\ -BIC$_{0.5}$ \end{tabular}} & FP & 45.44 & 35.83 & 32.36 & 31.63 & 30.23 & 28.71 & 27.60 & 25.94 \\ 
                       & FN & 296.12 & 245.04 & 192.77 & 151.15 & 114.42 & 86.13 & 65.16 & 47.90 \\ 
                       & HD & 341.56 & 280.87 & 225.13 & 182.78 & 144.65 & 114.84 & 92.76 & 73.84 \\
\midrule
\multirow{3}{*}{\begin{tabular}{@{}c} L1LR$_\land$ \\ -BIC$_{0.5}$ \end{tabular}} & FP & 4.03 & 2.55 & 1.67 & 1.42 & 1.12 & 0.69 & 0.54 & 0.48 \\
                       & FN & 340.89 & 291.34 & 239.26 & 195.04 & 152.79 & 119.74 & 94.39 & 71.29 \\
                       & HD & 344.92 & 293.89 & 240.93 & 196.46 & 153.91 & 120.43 & 94.93 & 71.77 \\
\midrule
\multirow{3}{*}{MPL} & FP & 15.94 & 7.53 & 3.05 & 0.96 & 0.49 & 0.15 & 0.08 & 0.03 \\
                       & FN & 311.71 & 265.97 & 219.21 & 179.76 & 144.11 & 114.19 & 90.80 & 70.23 \\
                       & HD & 327.65 & 273.50 & 222.26 & 180.72 & 144.60 & 114.34 & 90.88 & 70.26 \\
\midrule
\multirow{3}{*}{BJP} & FP & 0.10 & 0.11 & 0.22 & 0.38 & 0.87 & 1.20 & 1.73 & 2.08 \\
                       & FN & 435.67 & 408.55 & 366.81 & 318.43 & 283.71 & 236.68 & 208.95 & 186.30 \\ 
                       & HD & 435.77 & 408.66 & 367.03 & 318.81 & 284.58 & 237.88 & 210.68 & 188.38 \\ 
\multicolumn{10}{l}{} \\
\multicolumn{10}{c}{Hub network $(d = 256)$} \\
\multicolumn{2}{c}{Method \textbackslash $\frac{N}{1000}$} & .25 & .5 & 1 & 2 & 4 & 8 & 16 & 32 \\
\toprule
\multirow{3}{*}{\begin{tabular}{@{}c} PLRHC- \\ BIC$_{0.5}$ \end{tabular}} & FP & 19.40 & 13.19 &  8.01 &  5.30 &  3.78 &  2.54 &  1.67 &  1.08 \\ 
                       & FN & 160.34  & 132.91  & 104.33  &  80.23  &  61.17  &  44.80  &  34.10  &  23.63 \\ 
                       & HD & 179.74 & 146.10 & 112.34 &  85.53 &  64.95 &  47.34 &  35.77 &  24.71 \\
\midrule
\multirow{3}{*}{\begin{tabular}{@{}c} L1LR$_\lor$ \\ -BIC$_{0.5}$ \end{tabular}} & FP & 44.65 & 34.50 & 27.56 & 23.40 & 20.59 & 18.47 & 17.71 & 15.86 \\
                       & FN & 153.71 & 127.26 & 101.27 &  78.15 &  60.12 &  43.35 &  33.56 &  23.28 \\
                       & HD & 198.36 & 161.76 & 128.83 & 101.55 &  80.71 &  61.82 &  51.27 &  39.14 \\
\midrule
\multirow{3}{*}{\begin{tabular}{@{}c} L1LR$_\land$ \\ -BIC$_{0.5}$ \end{tabular}} & FP & 11.62 &  7.56 &  4.91 &  3.17 &  2.30 &  1.58 &  1.12 &  0.70 \\
                       & FN & 183.18 & 158.78 & 131.16 & 106.59 &  84.48 &  65.40 &  51.07 &  38.01 \\
                       & HD & 194.80 & 166.34 & 136.07 & 109.76 &  86.78 &  66.98 &  52.19 &  38.71 \\
\midrule
\multirow{3}{*}{MPL} & FP & 28.48 & 16.92 &  9.25 &  4.73 &  2.54 &  1.36 &  0.69 &  0.29 \\
                       & FN & 169.30 & 149.26 & 128.75 & 110.49 &  95.27 &  79.82 &  68.88 &  55.78 \\
                       & HD & 197.78 & 166.18 & 138.00 & 115.22 &  97.81 &  81.18 &  69.57 &  56.07 \\
\midrule
\multirow{3}{*}{BJP} & FP & 0.11 & 0.13 & 0.04 & 0.04 & 0.13 & 0.10 & 0.11 & 0.20 \\
                       & FN & 214.33 & 193.06 & 166.39 & 138.76 & 109.62 &  81.00 &  58.72 &  38.31 \\
                       & HD & 214.44 & 193.19 & 166.43 & 138.80 & 109.75 &  81.10 &  58.83 &  38.51 \\
\end{tabular}
}
\caption{Detailed results for the larger synthetic network experiments. FP = number of false positive edges, FN = number of false negative edges and HD = Hamming distance. Results are averaged over 100 sampled data sets.}
\label{table:comparison2}
\end{table}

\section{Discussion} \label{section:discussion}
In this work, we examined the problem of score-based learning of the graph structure of binary pairwise Markov networks. By utilizing the correlation decay property (Bresler, Mossel and Sly \citeyear{bresler2008}) to construct variable-wise correlation neighborhoods, we designed a greedy search algorithm, denoted PLRHC-BIC$_\gamma$, suitable for the pseudo-likelihood family of methods which scales better and is more accurate than contemporary methods. Under a common sparsity assumption, the number of candidates for each correlation neighborhood can be much smaller than the number of variables, which provides a significant reduction in the computational load. As the base algorithm, we considered the algorithm proposed by Pensar et al. (\citeyear{pensar2017}) which learns the structure in two phases, with the first phase being similar to the IAMB algorithm (Tsamardinos et al., \citeyear{tsamardinos2003}). We implemented the method for the BIC$_\gamma$ scoring function proposed by Barber and Drton(\citeyear{barber2015}) to enable a straightforward comparison against L1LR$_{\lor / \land}$-BIC$_\gamma$, proposed by the same authors. We note that the scoring function used in our method is exchangeable.

We showed that our approach, PLRHC-BIC$_\gamma$, performed favorably under all tested conditions against the L1LR$_{\lor / \land}$-BIC$_\gamma$ search algorithm. PLRHC-BIC$_\gamma$ had good performance for small as well as large sample sizes. It achieved a stable performance with both low false positive and false negative edge counts for all sample sizes, whereas L1LR$_{\lor / \land}$-BIC$_\gamma$ could, depending on a user-provided $\lor$ or $\land$ choice, achieve a low count in only one of them. We additionally showed that PLRHC-BIC$_\gamma$ performed favorably under all tested conditions against two other state-of-the-art Markov network structure learning algorithms.

While our approach can speed up pseudo-likelihood-based approaches significantly, constructing the correlation neighborhoods with the pairwise tests still requires a number of computations quadratic in the number of variables. To further improve the scalability of this family of methods, a natural direction for future work would be to develop even faster methods for constructing the candidate neighborhoods without sacrificing too much in terms of accuracy. We further note that while we restricted our study to binary pairwise networks, extending BIC$_\gamma$ to general discrete pairwise networks is straightforward. Additionally, another direction for extending the present work would be to generalize the scope of the method to general (non-pairwise) Markov networks by introducing interaction terms in the logistic regression model.

\bibliography{article}
\bibliographystyle{plainnat}

\end{document}